# Integrating Probabilistic Rules into Neural Networks: A Stochastic EM Learning Algorithm

Gerhard Paass*
International Computer Science Institute (ICSI)
1947 Center Street, Berkeley, California 94704
E-mail: paass@icsi.berkeley.edu

## Abstract

The EM-algorithm is a general procedure to get maximum likelihood estimates if part of the observations on the variables of a network are missing. In this paper a stochastic version of the algorithm is adapted to probabilistic neural networks describing the associative dependency of variables. These networks have a probability distribution, which is a special case of the distribution generated by probabilistic inference networks. Hence both types of networks can be combined allowing to integrate probabilistic rules as well as unspecified associations in a sound way. The resulting network may have a number of interesting features including cycles of probabilistic rules, hidden 'unobservable' variables, and uncertain and contradictory evidence.

## 1 INTRODUCTION

Probabilistic inference networks (Pearl 1988) have been used to model uncertain causal relations between variables, for instance in a diagnostic system. They consist of a number of rules each of which describes the probabilistic relation of few, typically two to five, variables. Each rule is assumed to model some sort of 'weak' causal dependency. Taken together these rules define the joint probability distribution of a large set of variables. Here we tacitly assume that according to the maximum entropy principle higher order interactions not affected by the rules are set to zero.

The rules should reflect theoretical or empirical knowledge about the corresponding domain. If, however, this knowledge is not available we may capture the probabilistic information in the data by an associative *neural network* (Anderson &

Rosenfeld 1988). Pairs of variables of such a network are connected by weighted 'links' modelling their 'correlation'. Its representational power is based upon additional artificial 'hidden' variables used to approximate higher order interactions. The unkown parameters (weights) of a network are automatically adapted to the data by estimation algorithms. Even complex dependencies can be approximated arbitrarily well if the number of hidden variables is sufficiently large (White 1989; Hertz et al. 1991, p.141ff).

The *Boltzmann machine* (Ackley et al. 1985) is a neural network which modifies its variables according to a joint probability distribution. Together with the probabilistic inference network it forms a structure which is able to represent the probabilistic rules as well as the associative data. An example of such a network describing the relation of discrete economic variables is shown in figure 1. All variables are discrete with the '+' meaning an increase of that quantity. The relation between some variables (Taxes+, Deficit+, Interest+, and Stocks+) is described by probabilistic rules based on theoretical considerations. To describe the relation involving the remaining variables (Taxes+, Employ.+, Product.+, and Stocks+) an associative neural network with hidden variables $H_1, \ldots, H_4$ is assumed, as no specific functional relations are known.

The associative data and the probabilistic rules in general will not be compatible. To arrive at a single joint distribution, we have to find some sort of compromise which is formed according to the relative reliability of the input information. The reliability of the information is described with a measurement distribution. The approach developed in this paper is able to combine conflicting information on probabilities and even may process networks with cycles.

The functional form of maximum entropy distributions of discrete variables subject to constraints has been derived twenty years ago by Darroch and Ratcliff (1972). Paass (1989) proposed the integration of neural networks and probabilistic inference networks using a variant of the Boltzmann machine learning algorithm. Recently Hrycej (1990) discussed the relation between Gibbs sampling and probabilistic reasoning. His approach, however, imposes some restrictions, e.g. acyclic network structure, complete knowledge of all conditional probabilities (no truncation), etc, which are not required in this paper.

---

*On leave from German National Research Institute for Computer Science (GMD), D-5205 St. Augustin; E-mail: paass@gmdzi.gmd.de

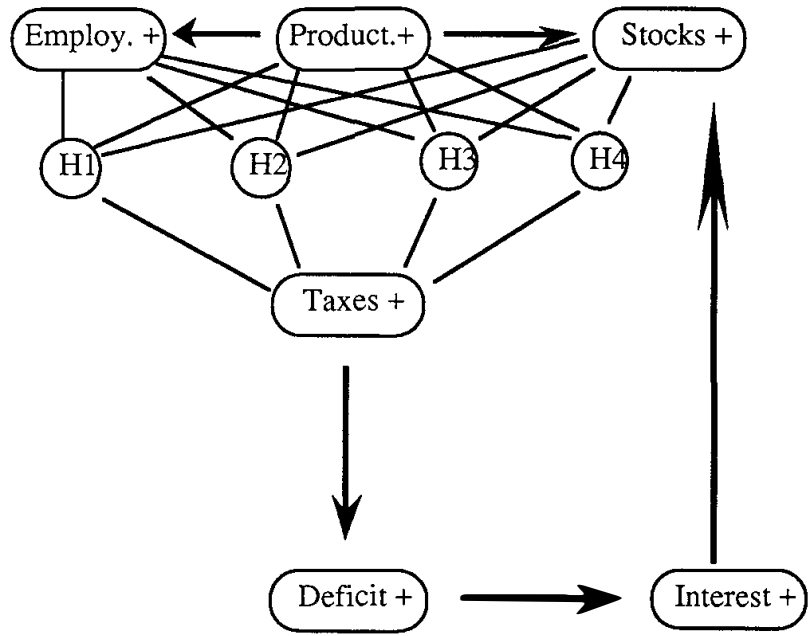

Figure 1: **Combined Network**

The next section contains the basic definitions of probabilistic inference networks and the related Boltzmann machine. In section three the maximum likelihood estimation from conflicting evidence is discussed. In section four a stochastic version of the EM-algorithm is used to determine the maximum likelihood estimate. The paper is concluded with a short discussion.

## 2  PROBABILISTIC NETWORKS

Consider a problem whose relevant features may completely be described in terms of $k$ atomic propositions $A_1, \ldots, A_k$. Corresponding to each $A_i$, a random variable $x_i$ is defined taking the values 1 if $A_i$ holds and 0 otherwise. The variables are collected in a vector $x := (x_1, \ldots, x_k)$ whose $2^k$ different values are called 'possible worlds' and form a set $\mathcal{X}$. If $\mathcal{B}$ is the Boolean algebra generated from the $A_i$, each proposition $B \in \mathcal{B}$ corresponds to the subset of $\mathcal{X}_B \subset \mathcal{X}$ where $B$ holds. To arrive at a simpler notation we write $x \in B$ instead of $x \in \mathcal{X}_B$. The available information on the probability of the propositions is compiled into a joint probability distribution $p : \mathcal{B} \to [l, \infty]$. The probability of some proposition $B \in \mathcal{B}$ is defined as $p(B) := p(\mathcal{X}_B) = \sum_{x \in B} p(x)$.

This setup is used for probabilistic inference networks as well as for associative neural networks. Therefore we can integrate both approaches using the common probability measure $p(x)$. The structure of $p(x)$ is assumed to be known in advance.

Let us first consider a *probabilistic inference network* where the expert's knowledge may be stated in terms of marginal probabilities, e.g. $p(C_r) = q_r$, as well as 'probabilistic rules' which may be interpreted as restrictions on conditional probabilities, e.g. $p(C_r \mid B_r) = q_r$ for propositions $C_r, B_r \in \mathcal{B}$. These restrictions can be reformulated in terms of linear constraints of the form

$$\sum_{x \in \mathcal{X}} b_r(x) p(x) = c_r \qquad r = 1, \ldots, d \qquad (1)$$

For marginal probabilities $p(C_r) = q_r$ the term $b_r(x)$ can be defined using the indicator function

$$b_r(x) := [C_r](x) := \begin{cases} 1 & \text{if } x \in C_r \\ 0 & \text{otherwise} \end{cases} \qquad (2)$$

and $c_r := p(C_r)$. For conditional probabilities $p(C_r \mid B_r) = q_r$ we set

$$b_r(x) := (1 - q_r)[C_r \wedge B_r](x) - q_r[\neg C_r \wedge B_r](x) \qquad (3)$$

and $c_r := 0$. If the constraints are not contradictory[1], there exists a probability distribution $p(x)$, where all of them hold simultaneously.

As in general the number $d$ of constraints is much lower than the number of all $2^k$ elementary probabilities $p(x)$, there is a large set $\mathcal{P}$ of different probability distributions, which simultaneously satisfy all constraints. Similar to (Cheeseman 1983) we select the distribution from $\mathcal{P}$ which *maximizes the*

---

[1] Later we are considering inconsistent constraints.

*entropy* $H(p) = -\sum_{x \in \mathcal{X}} p(x) \log p(x)$ subject to the equality constraints (1), as for this distribution lowest 'interactions' between the variables result. It is unique and has the functional form (Darroch & Ratcliff 1972)

$$p(x) = Z^{-1} \exp\left(\sum_{r=1}^{d} \lambda_r b_r(x)\right) \quad (4)$$

with a multiplicative constant

$$Z := \sum_{x \in \mathcal{X}} \exp\left(\sum_{r=1}^{d} \lambda_r b_r(x)\right)$$

which restricts the sum of probabilities to 1. The parameters $\lambda_r$ have to be determined in such a way that the constraints (1) hold.

If the parameters $\lambda_r$ are known, the equation (4) may be used to simulate the distribution $p(x)$ by successively generating new values for the variables exploiting

$$p(x_i=1 \mid \bar{x}_i) \quad (5)$$
$$= \frac{1}{1 + p(\bar{x}_{i0})/p(\bar{x}_{i1})}$$
$$= \frac{1}{1 + \exp\left(\sum_{r=1}^{d} \lambda_r \left[b_r(\bar{x}_{i0}) - b_r(\bar{x}_{i1})\right]\right)}$$

where $\bar{x}_i := (x_1, \ldots, x_{i-1}, x_{i+1}, \ldots, x_k)$ and $\bar{x}_{i1} := (x_1, \ldots, x_{i-1}, 1, x_{i+1}, \ldots, x_k)$. The terms $b_r(\bar{x}_{i1})$ and $b_r(\bar{x}_{i0})$ depend only on variables which are involved in restriction $r$ and there difference is zero if $x_i$ is not involved in that restriction. Then (5) shows that $p(x_i=1 \mid \bar{x}_i)$ is only dependent on the vector of varibles different from $x_i$, which simultaneously with $x_i$ are involved in some constraint (1). Such a structure is called a Markov random field (Kindermann & Snell 1980) and corresponds to a a *nearest neighbor Gibbs potential* (Paass 1989; Hrycej 1990). To generate values according to this distribution we may start with an arbitrary vector $x$, select components $x_i$ at random, and alter their values according to (5). The sequence of vectors evolving from this procedure will have the desired distribution.

The *Boltzmann machine* (Ackley et al. 1985) is a probabilistic *neural network* where some pairs of variables $x_{i_r}, x_{j_r}$, $r = 1, \ldots, d$, are connected by links with weights $\lambda_r \in \Re$ indicating the mutual dependency or 'correlation' of $x_{i_r}$ and $x_{j_r}$. Note that $\lambda_r = 0$ if there is no direct dependency. Using

$$b_r(x) := [A_i \wedge A_{j_r}](x) \quad (6)$$

the probability of a possible world $x$ is defined by (4) (Aarts & Korst 1988, p.207). Hence the Boltzmann machine generates a distribution that has the same form as the maximum entropy distribution subject to the restriction of $p(A_i \wedge A_{j_r})$, $r = 1, \ldots, d$, to some value. Therefore uncertain reasoning in probabilistic inference networks as well as 'associative reasoning' in neural networks may be combined within one framework. In a neural network some of the variables are *hidden units*, for whom there are no observations available. These hidden units have no simple symbolic interpretation. They are, however, capable to represent arbitrary probabilistic relations, whereas networks without hidden units may only represent linear dependencies. The representational capabilities of neural networks are discussed by Hertz et al. (1991, p.141ff).

## 3 MAXIMUM LIKELIHOOD ESTIMATION

The structure of the network has to be fixed in advance. Accordingly we have to determine which variables directly interact – as indicated in the example in figure 1. In addition we know the functional form (4) of the probabilities except for the numerical parameters. For each probabilistic inference rule there is a $b_r(x)$-term according to (1), while each bivariate link in an associative 'neural network' substructure corresponds to an appropriate $b_r(x)$-term according to (6).

The probability values assigned to rules and especially the associative data may be subject to some error and in general are contradictory. Therefore these data items, denoted by $\tilde{q}_r$, are *assumed* to originate from independent random samples $S_r$ with $n_r$ elements generated according to the true distribution. If $\tilde{q}_r$ corresponds to the probability of some proposition $p(C_r)$ then we do not know the values for all variables $x_i$, but we only know whether $C_r$ holds or not. The fraction of records where $C_r$ holds is just our observed probability $\tilde{q}_r$. Hence we have a *missing data* situation. We get the binomial distribution $P(\tilde{q}_r \mid q_{r,\lambda})$ as the 'sampling distribution' describing the deviation of the observed probability $\tilde{q}_r$ from the theoretical value $p(C_r)$. This deviation gets smaller with increasing sample size $n_r$. As our samples are imaginary[2] we can select $n_r$ in such a way that, for instance, the true probability $p(C_r)$ is contained in a given interval $[a, b]$ with a probability of, say, 0.9.

If $\tilde{q}_r$ corresponds to a probabilistic rule $p(C_r \mid B_r)$, the sample $S_r$ is generated in a twostep procedure. First a sample $\check{S}_r$ of size $N_r$ is selected from the complete distribution. Then all sample elements where $B_r$ does *not* hold are removed. For each element of the remaining sample $S_r$ of size $n_r$ it is only reported whether $C_r$ holds or not. As part of the population is ignored, $S_r$ is called a *truncated sample*. The deviation between $\tilde{q}_r$ and $p(C_r \mid B_r)$ again is described by a binomial distribution where

---

[2] In the case that real samples are available, eg. from a statistical survey or a measurement device, we may use them instead.

Integrating Probabilistic Rules into Neural Networks: A Stochastic EM Learning Algorithm     267

$n_r$ can be selected to reflect the reliability of the value.

Data on the associative relation between variables also can be understood as an independent sample $S_r$ of size $n_r$ covering a subvector $y$ of visible variables. This time the sample is assumed to stem from real observations of the system in question. There is no special constraint related to $S_r$ as the stochastic relation between the $y$-variables are communicated by hidden variables not contained in $y$. To illustrate the situation consider the following example. Assume we have $k = 5$ variables $x_1, \ldots, x_5$ and three pieces of information:

$S_1$: a sample with $n_1 = 20$ elements on the marginal probability $p(C_1)$ with $C_1 = \{x \mid x_1=1\}$ and an observed relative frequency $\tilde{q}_1 = 0.8$.

$S_2$: a sample with $n_2 = 10$ elements on the conditional probability $p(C_2 \mid B_2)$ with $C_2 = \{x \mid x_4=1\}$ and $B_2 = \{x \mid x_1=1 \wedge x_2=1\}$ and an observed relative frequency $\tilde{q}_2 = 0.3$.

$S_3$: a sample with $n_3 = 10$ elements on the stochastic relation between the variables $y = (x_2, x_3, x_4)$. To communicate this relation we have symmetric bivariate links between the hidden variable $x_5$ and the visible variables $x_2, x_3, x_4$.

We must relate these samples to the joint distribution. While $S_1$ and $S_3$ are assumed to cover the joint distribution, the sample $S_2$ is truncated to the subset where $B_2$ is valid. Therefore we use the 'extended' sample $\check{S}^2$ which contains also elements where $\neg B_2$ holds. However, the number $\bar{n}_2$ of these records is unknown. Indicating missing data items by '?' table 1 shows the resulting records in the samples. In the sample $\check{S}^2$ we even do not know the sample size $N_2 := n_2 + \bar{n}_2$, as part of the records are missing. We may pool together all these samples to a comprehensive sample $S$ which in turn may be as a random sample from our distribution. In our example it is defined as $S := (S_1, \check{S}^2, S_3)$. Note that there may exist several samples corresponding to different associative sub-networks. Similar to the conditional probabilities of rules these samples may be truncated, i.e. cover specific situations only.

Assuming that all information about the parameters of the distribution is contained in $S$ and that all samples have been obtained independently, we may use the maximum likelihood approach (cf. Paass 1988) to determine the optimal parameter $\hat{\lambda}$ as the solution of

$$\prod_r P(\tilde{q}_r \mid q_{r,\hat{\lambda}}) = \max_\lambda \prod_r P(\tilde{q}_r \mid q_{r,\lambda}) \quad (7)$$

In (Paass 1989) the derivatives of this likelihood function with respect to the parameters $\lambda_r$ are calculated. Starting with some parameter values we subsequently may use gradient techniques to determine the maximum. The resulting 'generalized Boltzmann machine learning algorithm' has the characteristic that for the current $\lambda$-values specific probabilities have to be estimated by stochastic simulation using (5). In addition a non-linear equation system has to be solved for each iteration if cycles are present in the network which involve probabilistic rules.

The Boltzmann machine is very computation intensive. Nevertheless in the area of associative networks they are found to capture the underlying statistical relations in a very effective way. In a detailed comparison on a statistical decision task, Kohonen et al. (1988) found that the Boltzmann machine achieved considerably better accuracy than a backpropagation network. The choice of various process parameters is an active research field (Hertz et al. 1991, p.168ff).

## 4  THE STOCHASTIC EM-ALGORITHM

As an alternative we consider a sample-based procedure to determine the parameters of $p(x)$. In essence we reconstructs the missing items of the pooled sample $S =: (x_{(1)}, \ldots, x_{(n)})$ described above, whose elements are denoted by $x_{(j)}$. This is just the approach of the stochastic EM-algorithm (Celeux & Diebolt 1988), which is a random version of a general procedure for handling missing data in maximum likelihood problems (Dempster et al. 1977). This algorithm starts with some arbitrary[3] parameter vector $\hat{\lambda}$ and iterates the following steps:

E-step:
    Assume $x_{(j)} = (y_{(j)}, z_{(j)})$ is an arbitrary record of the comprehensive sample $S$ and let $y_{(j)}$ be the vector of actually observed values. Then for each $y_{(j)}$ the value of $z_{(j)}$ is randomly generated according to the conditional distribution $p_{\hat{\lambda}}(z_{(j)} \mid y_{(j)})$ given the values $y_{(j)}$ and the current parameter $\hat{\lambda}$. In the case of truncated samples $\check{S}_i$ the expected value of the sample size $\bar{n}_i$ of the truncated portion is estimated. Hence all missing data items are replaced by imputed values.

M-step:
    In this step a maximum likelihood estimation of the parameters $\lambda$ is performed using the imputed values as if they were actually observed. With the new $\hat{\lambda}$ the E-step is performed again.

The procedure stops if the parameter vector $\hat{\lambda}$ reaches a stationary point. In some sense the sam-ple $S$ can be understood as a parametrization of the complete distribution. By the law of large numbers the approximation of the distribution gets better if the sample size $n$ is increased, for instance by duplicating each record in $S$ sufficiently often. For $n \to \infty$ the distribution can be represented arbitrarily well.

---
[3] The starting parameters should be different from saddlepoints, as the procedure stops there. For associative data this means that the hidden variables should be dependent on the visible variables, i.e $\lambda_r \neq 0$.

Table 1: **Evidence in the Form of Samples with Missing Values**

| Sample | Sample Size | No. of Records | \multicolumn{5}{c}{Values of Variables} |
|--------|-------------|----------------|-------|-------|-------|-------|-------|
|        |             |                | $x_1$ | $x_2$ | $x_3$ | $x_4$ | $x_5$ |
| $S_1$  | $n_1 = 20$  |                |       |       |       |       |       |
|        |             | 4              | 0     | ?     | ?     | ?     | ?     |
|        |             | 16             | 1     | ?     | ?     | ?     | ?     |
| $\bar{S}^2$ | $N_2 = 10 + \bar{n}_2$ |     |       |       |       |       |       |
|        |             | ?              | 0     | 0     | ?     | ?     | ?     |
|        |             | ?              | 0     | 1     | ?     | ?     | ?     |
|        |             | ?              | 1     | 0     | ?     | ?     | ?     |
|        |             | 7              | 1     | 1     | ?     | 0     | ?     |
|        |             | 3              | 1     | 1     | ?     | 1     | ?     |
| $S_3$  | $n_3 = 10$  |                |       |       |       |       |       |
|        |             | 1              | ?     | 0     | 0     | 0     | ?     |
|        |             | 2              | ?     | 1     | 0     | 0     | ?     |
|        |             | 2              | ?     | 1     | 0     | 1     | ?     |
|        |             | 4              | ?     | 1     | 1     | 0     | ?     |
|        |             | 1              | ?     | 1     | 1     | 1     | ?     |

It has been shown (Celeux & Diebolt 1988) that for $n \to \infty$ under rather general conditions the parameter $\hat{\lambda}$ estimated by the stochastic EM algorithm corresponds to a local minimum of the likelihood function. Empirical evidence shows that the stochastic imputation step allows the algorithm to escape from local minima. The convergence properties of the usual EM-algorithm are discussed by Wu (1983).

To perform the stochastic *E-step* we first have to estimate the weights of truntcated records in truncated samples. These weights simply are selected as the empirical fractions of records with the corresponding values in the current complete sample $S$. This gives the new sample sizes $N_i$ of the truncated samples. Then we may use (5) together with (4) to generate new values stochastically according to the current value of $\hat{\lambda}$. For each $x_{(j)}$ we start with the present values and randomly select a component of $z_{(j)}$. Its value is randomly determined using (5). After a number of such modifications $z_{(j)}$ fluctuates according to the distribution $p(z_{(j)} | y_{(j)})$. The adaption to the new distribution will be particularly fast as the existing values are used as starting states and the difference between the conditional distributions usually will be small.

For the *M-step* we know that for each variable $x_i$ the conditional probabilities $p(x_i | \bar{x}_i)$ should follow the relations (5) and (4). From the binomial distribution we get the log-likelihood function

$$L_i = \sum_{j=1}^{n} L_{ij} \qquad (8)$$

$$= \sum_{j=1}^{n} \sum_{m=0}^{1} \tilde{p}(x_{i(j)}=m | \bar{x}_{i(j)})$$

$$* \log p(x_{i(j)}=m | \bar{x}_{i(j)})$$

where $\tilde{p}(x_{i(j)}=1 | \bar{x}_{i(j)})$ is the observed probability in record $x_{(j)}$, i.e. has the value 0 or 1. The derivative of $L_i$ with respect to $\lambda_r$ is, using (5), given by

$$\frac{\partial L_{ij}}{\partial \lambda_r} = \left[ \frac{\tilde{p}(x_i=1 | \bar{x}_i)}{p(x_i=1 | \bar{x}_i)} - \frac{1 - \tilde{p}(x_i=1 | \bar{x}_i)}{1 - p(x_i=1 | \bar{x}_i)} \right]$$
$$* \frac{\partial p(x_i=1 | \bar{x}_i)}{\partial \lambda_r}$$

We have omitted the index $(j)$ for simplicity. Defining $\bar{x}_{i1} := (x_i=1, \bar{x}_i)$ and $R_i := p(x_i=0, \bar{x}_i)/p(x_i=1, \bar{x}_i)$ we find from (5)

$$\frac{\partial p(x_i=1 | \bar{x}_i)}{\partial \lambda_r} = \frac{-\frac{\partial R_i}{\partial \lambda_r}}{(1+R_i)^2} \qquad (9)$$

As $R_i = \exp\left(\sum_{s=1}^{d} \lambda_s [b_r(\bar{x}_{i0}) - b_r(\bar{x}_{i0})]\right)$ we get

$$\frac{\partial R_i}{\partial \lambda_r} = R_i [b_r(\bar{x}_{i0}) - b_r(\bar{x}_{i0})]$$

Note that $R_i$ can be determined from (5) and (4) using the current parameters $\lambda_s$. According to the Hammersley-Clifford theorem (Besag 1974) the distribution $p(x)$ is completely determined if we know the conditional distributions $p(x_i | \bar{x}_i)$, $i = 1, \ldots, k$. Hence the estimation of the conditional distributions from our sample completely determines the unkown parameters $\lambda$. We evaluate $\frac{\partial L_i}{\partial \lambda_r}$ for each $x_i$ and modify the current values of $\lambda_r$ according to $\sum_{i=1}^{k} \frac{\partial L_i}{\partial \lambda_r}$. The maximumum likelihood estimates are developed under the assumption that we have for each $x_i$ an independent version of the synthetic sample $S$. This can be generated by stochastic simulation. However, preliminary experience shows that we may use a single sample for all variables.

An alternative approach uses (4) to express the log-likelihood of the pooled sample $S$

$$\log p(S) = \sum_{j=1}^{n} \log p(x_{(j)}) \qquad (10)$$

$$= \sum_{j=1}^{n} \left( \sum_{s=1}^{d} \lambda_s b_s(x_{(j)}) \right) - \log(Z)$$

We get the derivatives

$$\frac{\partial \log p(x_{(j)})}{\partial \lambda_r} = b_r(x_{(j)}) - \frac{\partial \log(Z)}{\partial \lambda_r} \qquad (11)$$

$$\begin{aligned}
\frac{\partial \log Z}{\partial \lambda_r} &= \frac{1}{Z} \sum_{x \in \mathcal{X}} \exp\left( \sum_{s=1}^{d} \lambda_s b_s(x) \right) b_r(x) \\
&= \sum_{x \in \mathcal{X}} \frac{\exp\left( \sum_{s=1}^{d} \lambda_s b_s(x) \right)}{Z} b_r(x) \\
&= \sum_{x \in \mathcal{X}} p_\lambda(x) b_r(x) \\
&= E_\lambda(b_r)
\end{aligned}$$

which is just the expected value of $b_r(x)$ for the current $\lambda$. Hence the derivative with respect to $\lambda_r$

$$\frac{\partial \log p(S)}{\partial \lambda_r} = \sum_{j=1}^{n} [b_r(x_{(j)}) - E_\lambda(b_r)] \qquad (12)$$

is just the sum of differences between the mean value of $b_r$ for the distribution with parameter $\lambda$ and the actual values of $b_r(x_{(j)})$ for the elements of the sample. This is a simplified version of the Boltzmann machine learning algorithm.

## 5 DISCUSSION

Similar to the Boltzmann machine the stochastic EM-algorithm involves a stochastic simulation of the variables. As only missing data items have to be imputed we have a 'clamped' simulation where the values of variables are used if they are known. In contrast to (Paass 1989) it is not necessary to solve a nonlinear equation system for each iteration. Currently empirical investigations are carried out to determine the relative computational efficiency of the stochastic EM-approach.

The algorithm developed in this paper may be used to incorporate probabilistic rules in a 'soft' way. We may start with a neural network containing only bivariate links (6) and introduce probabilistic rules into a joint sample simply as data as shown in table 1. After the learning algorithm has adapted to the data it should be able reproduce the rules in an approximate way. But the network is only an approximation to the maximum entropy distribution (4) which results if the probabilistic rules are 'hardwired' into the network. The conditional independence assumptions implied by the structure of the rule network may be invalid to some degree. There is more research needed to judge the effect of such errors. The EM-algorithm allows to compare the 'soft' with the 'hardwired' approach in a unified way.

## Acknowledgements

This work was supported in part by the German Federal Department of Research and Technology, grant ITW8900A7.

## References

Aarts, E., Korst, J. (1988). *Simulated Annealing and Boltzmann Machines.* Wiley, Chichester

Ackley, D., Hinton, G.E., Sejnowski, T.J. (1985). A Learning Algorithm for the Boltzmann machine. *Cognitive Science,* Vol.9 pp.147-169

Anderson, J.A., Rosenfeld, E. (1988). *Neurocomputing: Foundations of Research.* MIT Press, Cambridge, Ma.

Besag, J. (1974). Spatial Interaction and Statistical Analysis of Lattice Systems. *Journal of The Royal Statistical Society,* Series B., p.192-236

Celeux, G., Diebolt, J. (1988). *A Random Imputation Principle: The Stochastic EM Algorithm.* Tech. Rep. No.901, INRIA, 78153 Le Chesnay, France

Cheeseman, P. (1983). A Method of Computing Generalized Bayesian Probability Values for Expert Systems. Proc. *IJCAI'83.* Kaufmann, Los Altos, California.

Darroch, J.N., Ratcliff, D. (1972). Generalized Iterative Scaling for Log-Linear Models. *The Annals of Mathematical Statistics,* Vol.43, p.1470-1480

Dempster, A.P., Laird, N.M., Rubin, D.B. (1977). Maximum Likelihood from Incomplete Data via the EM algorithm (with discussion). *Journal of the Royal Statistical Society,* Vol.B-39, p.1-38

Hrycej, T. (1990). Gibbs Sampling in Bayesian Networks. *Artificial Intelligence.* Vol. 46, p.351-363

Hertz, J., Krogh, A., Palmer, R.G. (1991). *Introduction to the Theory of Neural Computation.* Addison Wesley, Redwood City.

Kindermann, R., Snell, J.L. (1980). *Markov Random Fields and their Applications.* American Math. Society, Providence, R.I.

Kohonen, Barna, G., Chrisley, R. (1988). Statis-

tical Pattern Recognition with Neural Networks: Benchmarking Studies. In proc. *IEEE International Conference on Neural Networks* (San Diego 1988), vol. I, p. 61-68. New York: IEEE.

Paass, G. (1988). Probabilistic Logic. In: Smets, P., A. Mamdani, D.Dubois, H.Prade (eds.) *Non-Standard Logics for Automated Reasoning*, Academic Press, London, p.213-252

Paass, G. (1989). Structured Probabilistic Neural Networks. Proc. *Neuro-Nimes '89* p.345-359

Pearl, J. (1988). *Probabilistic Reasoning in Intelligent Systems*, Morgan Kaufmann, San Mateo, Cal.

White, H. (1989). Some Asymptotic Results for Learning in Single Layer Feedforward Network Models. *J. American Statistical Association.* Vol.84, p.1003-1013.

Wu, C.F. (1983). On the Convergence Properties of the EM algorithm. *Annals of Statistics.* Vol.11, p.95-103.